\documentclass[letterpaper]{article}
\usepackage{aaai20}
\usepackage{times} \usepackage{helvet} \usepackage{courier} \usepackage[hyphens]{url} \usepackage{graphicx}   \usepackage{graphicx}   

\usepackage{pbox}
\usepackage{subcaption}
\usepackage{ifxetex}
\usepackage{ifluatex}

\ifxetex
  \usepackage[ios,font=seguiemj.ttf]{emoji}
  \usepackage{fontspec}
\else
  \ifluatex
    \usepackage[ios,font=Symbola_hint.ttf]{emoji}
    \usepackage{fontspec}
  \else
    \usepackage[T1]{fontenc}
    \usepackage[utf8]{inputenc}
    \usepackage[ios]{emoji}
  \fi
\fi

\usepackage{comment}
\usepackage{pbox}
\usepackage{siunitx}
\usepackage{amsmath}
\usepackage{booktabs} %
\usepackage{url}
\usepackage{graphicx}
\usepackage{hyperref}
\usepackage{graphicx}
\usepackage{color, colortbl}
\usepackage{arydshln}
\usepackage{tikz}
\usepackage[utf8]{inputenc}
\usepackage{units}
\usepackage{enumitem}
\usepackage{balance}
\usepackage[export]{adjustbox}
\usepackage{multirow}

\def\permille{\ensuremath{{}^\text{o}\mkern-5mu/\mkern-3mu_\text{oo}}}

\newcommand{\ie}{i.e.,\,}
\newcommand{\Ie}{I.e.,\,}
\newcommand{\eg}{e.g.,\,}
\newcommand{\Eg}{E.g.,\,}
\newcommand{\etal}{et~al\@ifnextchar.{}{.\@}}
\newcommand{\etc}{etc\@ifnextchar.{}{.\@}}

\definecolor{Gray}{gray}{0.9}

\newcommand{\afblock}[1]{\noindent{\textbf{#1 }}}

\title{Word-Emoji Embeddings from large scale Messaging Data reflect real-world Semantic Associations of Expressive Icons\thanks{To appear in: 3rd International Workshop on Emoji Understanding and Applications in Social Media}}

\author{
Jens Helge Reelfs\\
Brandenburg University\\of Technology\\
reelfs@b-tu.de
\And
Oliver Hohlfeld\\
Brandenburg University\\of Technology\\
hohlfeld@b-tu.de
\And
Markus Strohmaier\\
RWTH Aachen University\\\& GESIS\\
markus.strohmaier@cssh.rwth-aachen.de\\
\And
Niklas Henckell\\
The Jodel Venture GmbH\\
niklas@jodel.com
}

\makeatletter\ifdefined\Hy@StartlinkName
\def\addOneNestingLevelStartLink{%
  \gdef\Hy@StartlinkName##1##2{%
    \sbox0{\Hy@StartlinkNameOrig{##1}{##2}}\usebox0
    \global\let\Hy@StartlinkName\Hy@StartlinkNameOrig%
  }%
}
\def\addOneNestingLevelEndLink{%
  \gdef\pdfendlink{%
    \sbox0{\pdfendlinkOrig}\usebox0%
    \global\let\pdfendlink\pdfendlinkOrig%
  }%
}
\let\Hy@StartlinkNameOrig\Hy@StartlinkName
\let\pdfendlinkOrig\pdfendlink
\else
\let\addOneNestingLevelStartLink\relax
\let\addOneNestingLevelEndLink\relax
\fi

\begin{document}
\maketitle

\begin{abstract}
We train word-emoji embeddings on large scale messaging data obtained from the Jodel online social network. 
Our data set contains more than 40 million sentences, of which 11 million sentences are annotated with a subset of the Unicode 13.0 standard Emoji list. 
We explore semantic emoji associations contained in this embedding by analyzing associations between emojis, between emojis and text, and between text and emojis. 
Our investigations demonstrate anecdotally that word-emoji embeddings trained on large scale messaging data can reflect real-world semantic associations. 
To enable further research we release the Jodel Emoji Embedding Dataset (JEED1488) containing 1488 emojis and their embeddings along 300 dimensions. 
\end{abstract}

\section{Introduction}

\begin{figure*}[ht]
    \centering
    \includegraphics[width=\linewidth]{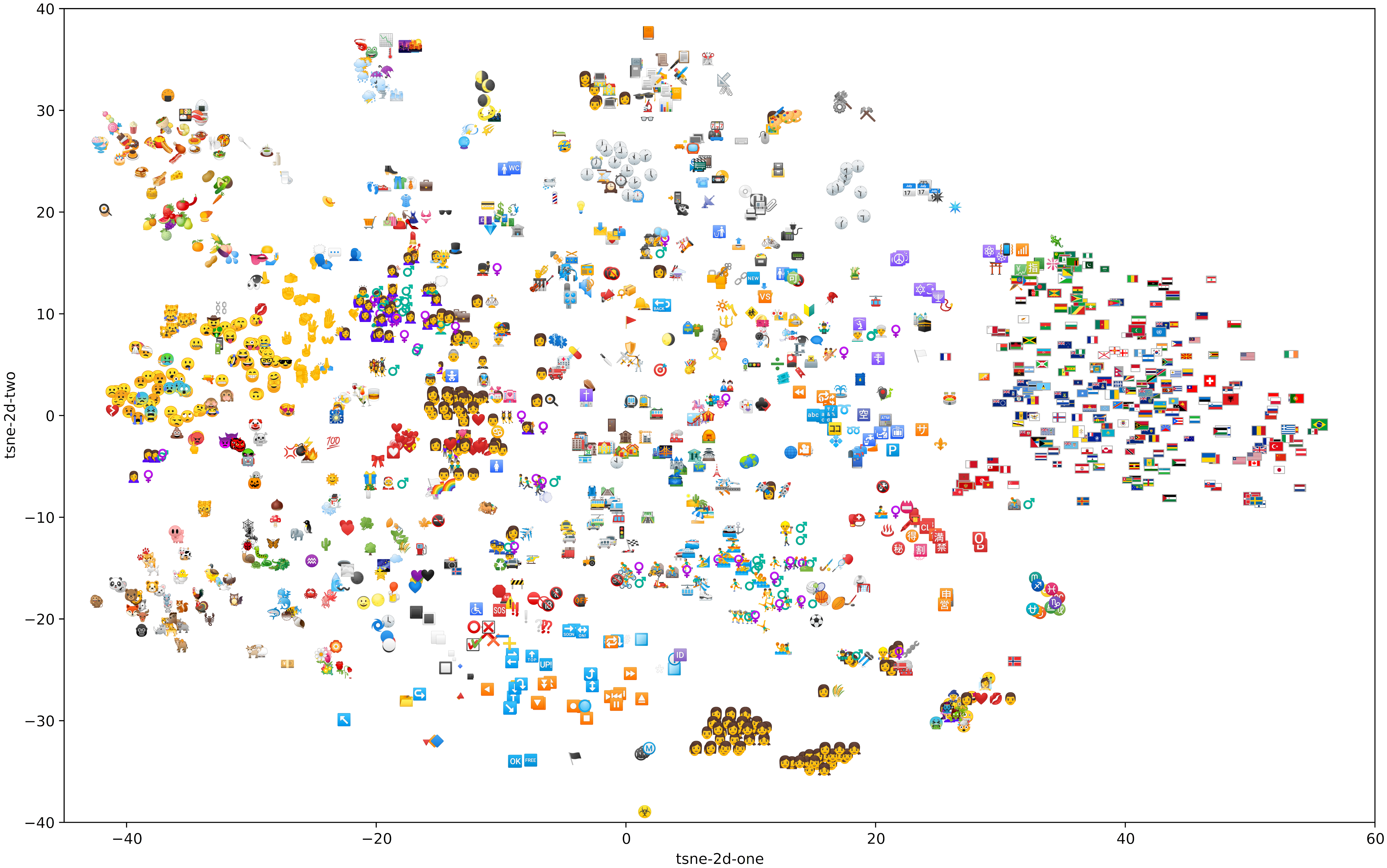}

    \caption{
        \textbf{Emoji2Emoji map.} It shows emoji associations as a 2D projection by their corresponding embedding vectors using the t-SNE algorithm.
        The figure highlights well formed semantic-based clusters of emoji associations: e.g., food (-40, 25) and fruits (-35, 20), animals (-35, -20), negative emotions faces (-40, 0), positive emotion faces (-30, 5), love, marriage, babies and families (-15, 0), country flags (50, 0), weather (-20, 35), or various alike signs (-5, -30).
        We observe that the embedding can uncover semantic relations between emoji from social network postings.
    }
    \label{fig:tsne}
\end{figure*}

We use four years of complete data from the online social network provider Jodel to explore semantic associations of emojis, \ie{} expressive icons, embedded in textual messaging data.
We deploy Word2Vec to generate combined word-Emoji embeddings and use the obtained embeddings to gauge the ability of Word-Emoji embeddings to capture different kinds of semantic emoji associations. In particular, we investigate:

\textbf{Associations between emojis:} What kind of associations between emojis are reflected by Word-Emoji embeddings of messaging data? We explore t-SNE projections of emoji associations (Figure~\ref{fig:tsne}) and interpret them qualitatively.
Our results show that Emoji2Emoji embeddings reveal insightful semantic associations beyond the Unicode standard.
    
\textbf{Associations between emojis and text:} What words are associated with a given emoji in a Word-Emoji embedding of messaging data? We explore the textual semantics of emojis by deriving top $k$ words that are most similar to a given emoji.
Our results highlight that quality emoji to text translations can be obtained from embeddings, e.g., to improve typing prediction on mobile devices or to inform social network users on emoji meanings in their network.
    
\textbf{Associations between text and emojis:} What emojis are associated with a given word in a Word-Emoji embedding of messaging data? We train machine learning models to predict an emoji for a given word and evaluate our results employing k-fold cross-validation.
Our results show that machine learning improves accuracy compared to a naive direct embedding approach at the cost of additional training.

These associations reflected by Word-Emoji embeddings trained on large scale message data open up a range of interesting downstream tasks and prospects, such as text to emoji translations, or emoji recommendation and replacement.

\begin{figure}[t]
    \centering
    \includegraphics[width=.75\linewidth]{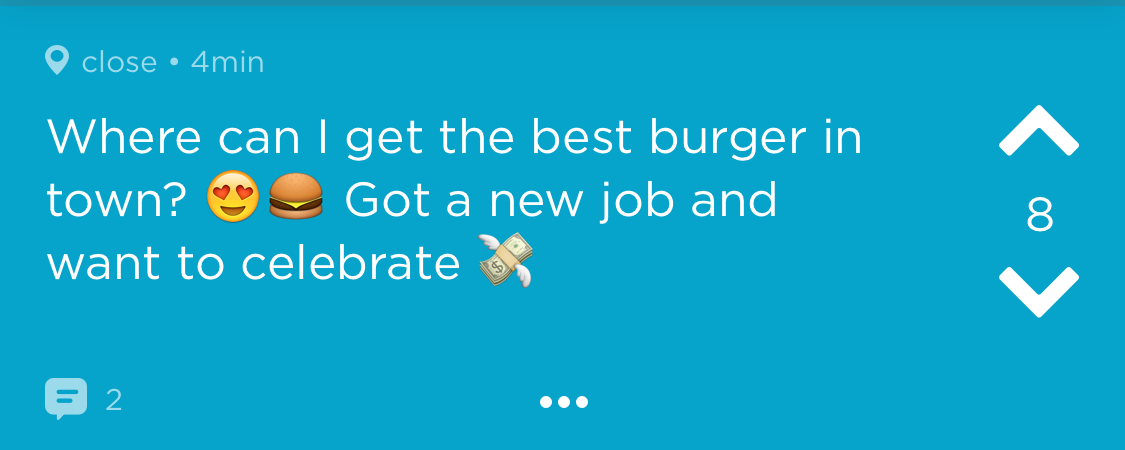}
    \caption{
        \textbf{Jodel example.}
        The Jodel messaging app supports posting short messages containing text and emojis in a user's geographical proximity.
    }
    \label{fig:jodel}
\end{figure}

\afblock{The Jodel Emoji Embedding Dataset (JEED1488).}
To enable further research, we release a subset of our embeddings to encourage and support further research into real-world semantic emoji associations.
This Jodel Emoji Embedding Dataset\footnote{\url{https://github.com/tuedelue/jeed}} containing 1488 emojis and their embedding along 300 dimensions based on word-emoji co-occurrence in a large messaging corpus.

\section{The Jodel Network \& Dataset}
    \label{sec:Dataset_Description_and_Statistics}
	
\afblock{The Jodel Network.}
    We base our study on a country-wide complete dataset of posts in the online social network Jodel\footnote{Jodel, German for yodeling, a form of singing or calling.}
    being a mobile-only messaging application. %
	It is location-based and establishes local communities relative to the users' location. %
	Within these communities, users can {\em anonymously} post a fotos from the camera app or content of up to 250 characters length, \ie{} microblogging, and reply to posts forming discussion threads.
	Posted content is referred to as {\em Jodels}. %
	Posts are only displayed to other users within close geographic proximity. %
	Further, all communication is {\em anonymous} to other users since no user handles nor other user-related information are shown.

Relative to other microblogging platforms and social networks, Jodel offers inherently different communication properties: due to its anonymity, there are no fixed user relations, nor profiles, nor any public social credit.
Unlike Twitter that has been found to be rather a news network with many Tweets sharing URLs~\cite{Haewoon2010whatistwitter}, URL posting is not enabled in Jodel embracing social exchange rather than news.
The location-based nature of Jodel can cause information to spread epidemically in-network, or triggered by external stimuli~\cite{reelfs2019jodelhashtags}.
In contrast to other social media channels, where statements are made to produce reactions and users promote themselves and compete for followers, Jodel seems to be driven to a considerable degree by posting questions and getting answers~\cite{fest-etal-2019-determining} subject to a community-based language regulation~\cite{heuman} in which local language policies are negotiated by community-members. 
This focus on social exchanges in local communities make Jodel an interesting platform to study emoji associations.

\begin{table}[t]
	\centering
	\small
	\begin{tabular}{|l|r|}
		\hline
		\textbf{Description}    & \textbf{\#}\\ \hline\hline
		Sentences			    & 48,804,375\\\hline
		{\raise.17ex\hbox{$\scriptstyle\mathtt{\sim}$}} 
		    after cleaning          & 42,245,200\\\hline
		Total Emojis            & 19,911,632\\\hline
		Sentences w/ Emojis     & 11,283,180\\\hline
	    Emojis per Sentence     & 1.76\\\hline
	    Unique Emojis w/ modifiers           & 2,620\\\hline
	    Uqunie Emojis w/o skin color       & 1,488\\\hline
	\end{tabular}
	\caption{
		\textbf{Dataset statistics.}
		The data set contains all posts in Germany since late 2014 until August 2017.
	}
	\label{tab:dataset}
\end{table}

	\afblock{Dataset.}
	The Jodel network operators provided us with data of their network, cf. Table~\ref{tab:dataset}.
	It contains posts content created within Germany only and spans multiple years from 2014 to 2017.
	Due to the Unicode Emoji standard developing over time, the dataset contains emojis from multiple versions and introduces a natural bias in frequency to elderly defined emojis; Emojis are available to the user via the standard OS' keyboard.
	That is, emoji interpretation might differ between users due to different pictogram representations~\cite{miller2016blissfully}; however, besides only being available to iOS and Android, different to the study from 2016, pictograms have harmonized to some extent.
	
    Note: Although $\approx13\%$ of all posts are images, we had to drop these posts due to lack of textual content data; within given figures, these posts are already excluded.

    \afblock{Ethics.}
	The dataset contains no personal information and cannot be used to personally identify users except for data that they willingly have posted on the platform.
	We inform and synchronize with the Jodel operator on analyses we perform on their data.

\section{Approach: Word-Emoji Embeddings}
    \label{sec_embedding}
    Word embeddings are semantic vector models that map written language information into an $n$-dimensional vector space representation.
    They have become a popular tool for both, industry and academia, \eg{} for finding word associations or sentiment analysis.

    Beyond classical embedding of words, emojis can likewise be incorporated in this process~\cite{eisner-etal-2016-emoji2vec,wijeratne2017emojinet}.
    Different off the shelf approaches exist to create word embeddings (\eg{} GloVe, fasttext, or Word2Vec) of which we use the gensim Word2Vec implementation due to its simplicity and popularity; the algorithm was proposed in ~\cite{mikolov2013efficient}.
    Word embedding is thus a promising approach to study emoji associations on Jodel.
    
    Our approach involves two steps: {\em i)} data preprocessing and cleaning, and {\em ii)} creating the embedding.

\afblock{Step 1: data preprocessing \& cleaning.}
    First, we dropped any skin color modifier to predict the type of emoji only.
    While preserving all emojis, we next cleaned the data by applying stemming and lemmatization using spaCy (built-in German model) and filtered out any non-alpha characters.
    We discard words less than 3 characters.
    Sentences consisting of less than two words were also discarded as they do not convey any information for our word embedding later.
    This resulted in $42,2\si{M}$ usable sentences.

\afblock{Step 2: embedding creation.}
    We create the word embedding by applying the gensim Word2Vec library to the entire corpus of usable sentences.
    Beyond default settings, we apply higher sampling to decrease the weight of most frequent words.
    Then, we trained the embedding model for 30 epochs; note that we discuss the variation of training in the next Section-{\em Varying the number of epochs}.

Next, we show that the resulting embeddings can be applied to accomplish three associative downstream tasks: {\em Emoji2Emoji}, {\em Emoji2Text}, and {\em Text2Emoji}.

\section{Emoji2Emoji Associations}
\label{sec:E2E}
    We begin by analyzing how emoji are semantically associated with each other on Jodel.
    That is, can we explain an emoji purely using other emojis.
    These associations can help social-network users to understand the \textit{subjective} meanings of an emoji in their local user community.

    \afblock{2D view on emoji associations.}
    We show the Emoji2Emoji associations as a 2D projection of all emoji by their embedding vectors using a t-SNE visualization in Figure~\ref{fig:tsne}~\cite{maaten2008visualizing}.
    The t-SNE algorithm iteratively clusters objects within a high dimensional vector space according to their distance and projects them to the selected number of dimensions---here 2D.
    The shown Figure is a hand-picked version of multiple projections with different random seeds for projection.
    Note that the rendering of complex emojis may be executed sequentially, \eg{} \emoji{emoji_u1f468_200d_2764_200d_1f48b_200d_1f468} is represented as \emoji{emoji_u1f469}\emoji{emoji_u2764}\emoji{emoji_u1f468}, or the profession \emoji{emoji_u1f469_200d_1f33e} is mapped to \emoji{emoji_u1f469}\emoji{emoji_u1f33e}. %
    
    We observe semantic-based clusters of emoji associations as they occur on Jodel.
    The distance within the projection often {\em can} be interpreted as semantic proximity, \eg{} there exist groups of almost all facial emoji (-35, 5).
    To mention few other examples, clusters (and their coordinates) involve food (-40, 25) and fruits (-35, 20), animals (-35, -20), negative emotions faces (-40, 0), positive emotion faces (-30, 5), families (-15, 0), country flags (50, 0), weather (-20, 35), or various alike signs (-5, -30).
    
    Next to the cluster of families (\emoji{emoji_u1f491}, \emoji{emoji_u1f46a}) (-15, 0), we find love (\emoji{emoji_u1f496}) (-20, -5), LGBT (\emoji{emoji_u1f3f3_200d_1f308}, \emoji{emoji_u1f46c}) (-15, -5), marriage (\emoji{emoji_u1f470}, \emoji{emoji_u1f48d}, \emoji{emoji_u1f492}) (-10, 5), pregnancy and children (\emoji{emoji_u1f930}, \emoji{emoji_u1f476}, \emoji{emoji_u1f6bc}) and even grandparents (\emoji{emoji_u1f475}, \emoji{emoji_u1f474}) next to it.
    
    For another example, the cluster of sports emojis (e.g., \emoji{emoji_u1f3c0}, \emoji{emoji_u1f3cc}) (10, -15) and especially water related sports (\emoji{emoji_u1f6a3_200d_2642}, \emoji{emoji_u1f3ca}, \emoji{emoji_u1f93d}) show that {\em holidays} (\emoji{emoji_u26f1}) may also closely be associated with flights (\emoji{emoji_u1f6ec}, \emoji{emoji_u1f6e9}), sea-born emojis \emoji{emoji_u1f6a2}, the globe (\emoji{emoji_u1f30e}, \emoji{emoji_u1f30f}), other attractions (\emoji{emoji_u1f3a2}, \emoji{emoji_u1f5fd}) and adventures (\emoji{emoji_u26fa}, \emoji{emoji_u1f5fb}).
    Further, next to any sort of biking (\emoji{emoji_u1f6b4}) as a sport, we observe transportation related emojis (\emoji{emoji_u1f6f5}, \emoji{emoji_u1f68e}) that may also indicate holidays.
    
    This represents a rich set of semantic associations derived from millions of posts reflecting real-world emoji semantics and there are lots of details to be discovered in each tSNE projection instance.
    However, while these associations are promising, the involved dimensional reduction {\em may} oversimplify the underlying vector space, consequently, it {\em may} pretend semantic associations that do not exist.
    Thus, we next take a deeper look into emoji similarity by analyzing their distances within the embedding space.
    
    \afblock{Emoji similarity.}
    To analyze emoji similarity, we first present a selection of emojis from different emoji groups (according to the Unicode definition) and their top $k$ most similar emojis from the embedding having a document frequency above 100 in Table~\ref{tab:emoji_similarity_example}.

    Most semantic associations are matching quite good, \eg{} \emoji{emoji_u1f37b} to other drinks, good vibes and party; \emoji{emoji_u1f3d5} to a tent, traveling and outdoor activities; \emoji{emoji_u1f609} to other positive emotions; etc.
    However, some symbols may have several semantic associations: while \emoji{emoji_u1f528} is related to other tools, the German word for hammer may also be used in explicit language as a synonym for the alternate meaning of \emoji{emoji_u1f346} as a phallic symbol.
    Another example can be seen in \emoji{emoji_u1f3f0}, which relates to mountains and attractions, but to presumably Harry Potter \emoji{emoji_u1f3eb} as well.
    The \emoji{emoji_u1f351} is mostly matched with other fruits.
    However, the association to \emoji{emoji_u1f346} relates to the mentioned alternative usage of both as symbols unrelated to the actual fruits, which may be in line with \emoji{emoji_u1f445}, \emoji{emoji_u1f4a6} and \emoji{emoji_u1f444}.
    In fact, the top 20 set contains more possibly fruit-unrelated emojis such as \emoji{emoji_u1f440}, \emoji{emoji_u1f528} and \emoji{emoji_u1f924}.
    Other fruits in this set (\emoji{emoji_u1f34e}, \emoji{emoji_u1f349}, \emoji{emoji_u1f345} and \emoji{emoji_u1f34a}) may thus be reinterpreted in a different context as well; Anyhow, this is a good example where the raw embedding is not well suited to distinguish between multiple semantics for a single item.

    \begin{table}[t]
        \centering
        \footnotesize
        \resizebox{.99\columnwidth}!{\begin{tabular}{c|cccccccccc}
            Emoji & 1 & 2 & 3 & 4 & 5 & 6 & 7 & 8 & 9 & 10 \\ \hline\hline
\emoji{emoji_u1f37b} & \emoji{emoji_u1f389} & \emoji{emoji_u1f37e} & \emoji{emoji_u1f37a} & \emoji{emoji_u1f64c} & \emoji{emoji_u1f377} & \emoji{emoji_u1f378} & \emoji{emoji_u1f44c} & \emoji{emoji_u1f60e} & \emoji{emoji_u1f601} & \emoji{emoji_u1f4aa}\\ \hline
\emoji{emoji_u1f3d5} & \emoji{emoji_u26fa} & \emoji{emoji_u1f3de} & \emoji{emoji_u1f5fa} & \emoji{emoji_u1f3c4} & \emoji{emoji_u1f3dd} & \emoji{emoji_u1f6a3} & \emoji{emoji_u1f3d9} & \emoji{emoji_u1f3e1} & \emoji{emoji_u1f5fd} & \emoji{emoji_u1f3a1}\\ \hline
\emoji{emoji_u1f575} & \emoji{emoji_u1f575_200d_2640} & \emoji{emoji_u1f440} & \emoji{emoji_u1f50e} & \emoji{emoji_u1f50d} & \emoji{emoji_u1f913} & \emoji{emoji_u1f46e} & \emoji{emoji_u261d} & \emoji{emoji_u1f446} & \emoji{emoji_u1f60e} & \emoji{emoji_u1f4b0}\\ \hline
\emoji{emoji_u1f308} & \emoji{emoji_u1f3f3_200d_1f308} & \emoji{emoji_u1f984} & \emoji{emoji_u2728} & \emoji{emoji_u1f49c} & \emoji{emoji_u1f49a} & \emoji{emoji_u1f60a} & \emoji{emoji_u1f31e} & \emoji{emoji_u1f46c} & \emoji{emoji_u2600} & \emoji{emoji_u1f338}\\ \hline
\emoji{emoji_u1f351} & \emoji{emoji_u1f346} & \emoji{emoji_u1f34c} & \emoji{emoji_u1f445} & \emoji{emoji_u1f4a6} & \emoji{emoji_u1f352} & \emoji{emoji_u1f60f} & \emoji{emoji_u1f449} & \emoji{emoji_u1f33d} & \emoji{emoji_u1f34a} & \emoji{emoji_u1f444}\\ \hline
\emoji{emoji_u1f528} & \emoji{emoji_u1f346} & \emoji{emoji_u1f6e0} & \emoji{emoji_u1f529} & \emoji{emoji_u1f449} & \emoji{emoji_u1f527} & \emoji{emoji_u1f44a} & \emoji{emoji_u1f4a5} & \emoji{emoji_u270a} & \emoji{emoji_u1f60f} & \emoji{emoji_u1f448}\\ \hline
\emoji{emoji_u1f3f0} & \emoji{emoji_u1f3de} & \emoji{emoji_u1f5fd} & \emoji{emoji_u1f3e1} & \emoji{emoji_u1f3eb} & \emoji{emoji_u1f6a2} & \emoji{emoji_u1f303} & \emoji{emoji_u1f5fb} & \emoji{emoji_u1f478} & \emoji{emoji_u1f3d8} & \emoji{emoji_u1f451}\\ \hline
\emoji{emoji_u1f648} & \emoji{emoji_u1f605} & \emoji{emoji_u1f644} & \emoji{emoji_u1f633} & \emoji{emoji_u1f64a} & \emoji{emoji_u1f604} & \emoji{emoji_u1f601} & \emoji{emoji_u1f602} & \emoji{emoji_u1f914} & \emoji{emoji_u1f629} & \emoji{emoji_u1f62c}\\ \hline
\emoji{emoji_u2795} & \emoji{emoji_u2716} & \emoji{emoji_u2796} & \emoji{emoji_u27a1} & \emoji{emoji_u1f6ab} & \emoji{emoji_u274c} & \emoji{emoji_u1f195} & \emoji{emoji_u25b6} & \emoji{emoji_u1f518} & \emoji{emoji_u1f51d} & \emoji{emoji_u2753}\\ \hline
\emoji{emoji_u1f609} & \emoji{emoji_u1f61c} & \emoji{emoji_u270c} & \emoji{emoji_u1f60b} & \emoji{emoji_u1f601} & \emoji{emoji_u1f603} & \emoji{emoji_u1f600} & \emoji{emoji_u1f607} & \emoji{emoji_u1f606} & \emoji{emoji_u1f618} & \emoji{emoji_u1f60f}\\ \hline
\emoji{emoji_u1f68c} & \emoji{emoji_u1f68d} & \emoji{emoji_u1f697} & \emoji{emoji_u1f683} & \emoji{emoji_u1f68e} & \emoji{emoji_u1f3ce} & \emoji{emoji_u1f4a8} & \emoji{emoji_u1f68b} & \emoji{emoji_u1f699} & \emoji{emoji_u1f6b4} & \emoji{emoji_u1f6b2}\\ \hline

        \end{tabular}
        }
        \caption{
            \textbf{Emoji top 10 similarity examples.} 
            We show a hand-picked selection of emojis and their closest emojis according to their vector cosine distance in our embedding.
            For most emojis, we observe good semantic relations.
            However, some emojis may have multiple semantics (castle---lock, home, traveling) or no apparent topic (plus sign).
            }
        \label{tab:emoji_similarity_example}
    \end{table}
    
\begin{figure}[t]
        \centering
        \includegraphics[width=\linewidth]{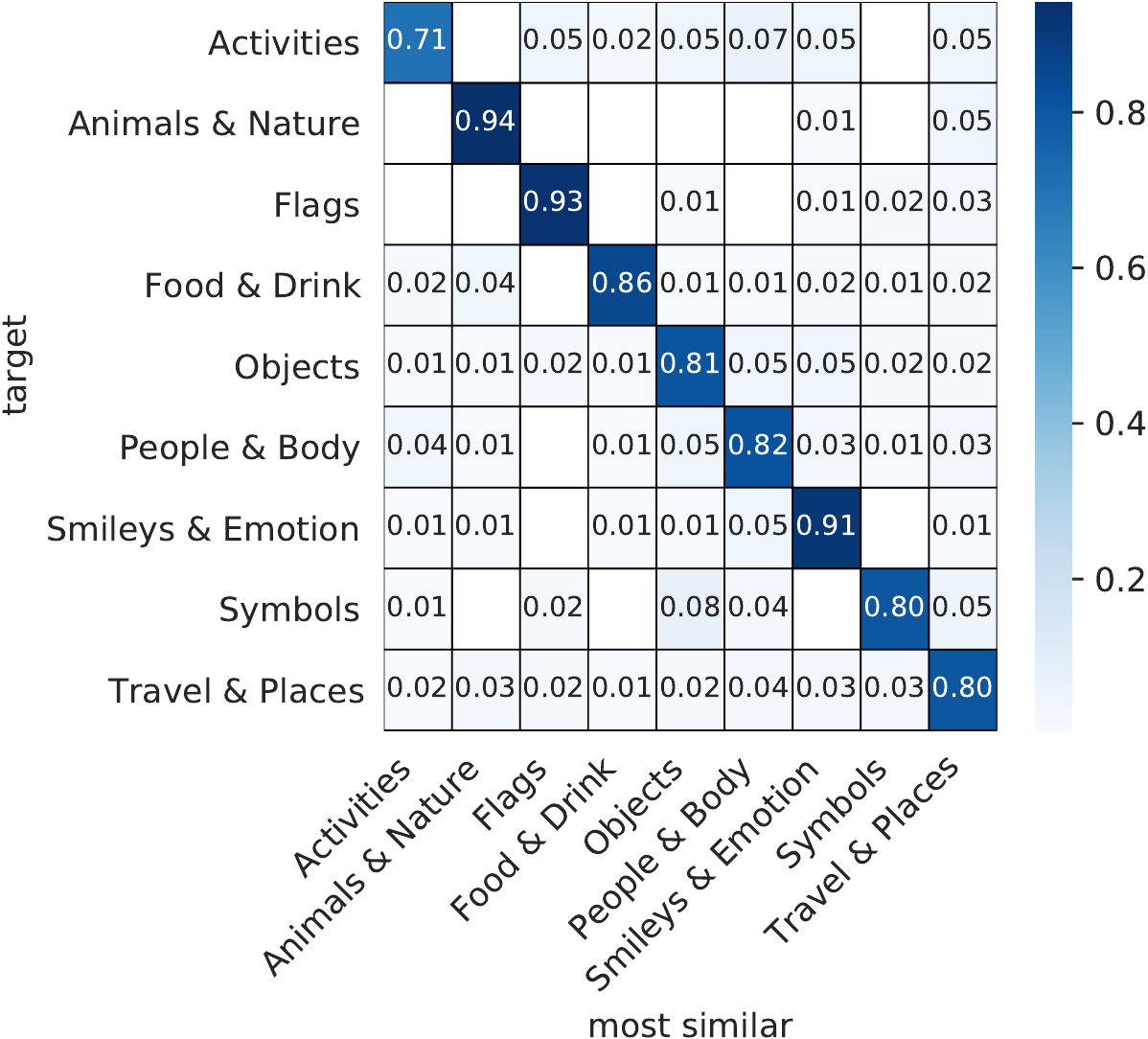}
        \caption{
            \textbf{Emoji2Emoji Unicode group association confusion matrix.}
            This heatmap shows emoji top1 associations from our embedding aggregated to their Unicode group as a confusion matrix.
            The values are normed row-wise.
            We observe an overall good similarity for the groups from our embedding depicted by the strong diagonal line.
        }
        \label{fig:emoji_similarity_groups}
    \end{figure}

    \afblock{Aggregation by Unicode groups.}
    To better understand these similarities, we next aggregate emoji into their \mbox{(sub-)}group according to the Unicode definition.
    Therefore, we show the confusion matrix of the topmost similar emoji pairs aggregated into their groups in Figure~\ref{fig:emoji_similarity_groups}.
    That is, we calculate the most similar emoji to each emoji contained in the dataset.
    Then, we count the in-/ correct mappings whether the most similar emoji's (x-axis, most similar) group is equal to the group of the compared emoji (y-axis, target).
    Due to the imbalance within the number of available emojis in each group, we further normalized the mappings to the total number of results emojis per group.
    
    We observe a strong straight diagonal indicating that most emoji are associated with other emoji in the same Unicode group (average precision of $0.8466 \pm 0.0932$).
    Deviations can mostly be explained by associated emoji located in different Unicode (sub-)groups (most notable for the Activities subgroup that has the amongst the lowest similarity scores).
    Example associations between different groups (noted in parentheses) include \emoji{emoji_u1f384} (Activities) to \emoji{emoji_u1f385} (person), \emoji{emoji_u2728} (Activities) to \emoji{emoji_u1f31f} (Animals \& Nature), \emoji{emoji_u1f380} (Activities) to \emoji{emoji_u1f49d} (Smileys \& Emotion), \emoji{emoji_u1f397} (Activities) to \emoji{emoji_u26b1} (Objects), \emoji{emoji_u26bd} (Activities) to \emoji{emoji_u1f1e9_1f1ea} (Flags).
    These results show that emoji embeddings on social network posts can reveal semantic associations beyond those captured by Unicode groups.
    
    Repeating this evaluation on a subgroup level results in the same observation (precision of $0.5918 \pm~0.3152$).
    Not surprisingly as seen within the t-SNE visualization in Figure~\ref{fig:tsne}, the main driver of confusion are the facial emojis in different subgroups.

    \afblock{Comparing embedding similarities with human judgment.}
    As last step, we compare our embedding and obtained emoji similarities to human perception leveraging the EmoSim 508 dataset from~\cite{wijeratne2017semantics}.
    It was obtained by asking 10 subjects to provide a similarity score for 508 well-selected different emoji pairs.
    To evaluate the suitability of an alike emoji-embedding, they apply the spearman rank correlation test achieving significant correlations for different model instances between $0.46$ and $0.76$.

    Applying the very same test to our embedding, our instance resulted in a moderate correlation of $0.5349$ with a very high significance, which is in line with Wijeratne et al. presented results.
    Anyhow, the asked user-base is not platform-specific to Jodel mismatching the used embeddings, which may make this comparison less representative.
    Eventually, by adjusting model parameters, this value can likely be improved.

    \afblock{Varying the number of epochs.} 
    While common wisdom may suggest {\em more is better}, our evaluation shows that the number of trained epochs impacts the results.
    By comparing the Emoji2Emoji and especially Emojis2Text results from embeddings that have been trained different numbers of epochs (5..100), we observed a negative shift in perceived capability to reflect \emph{multiple semantics} / synonyms with more training.
    \Ie{} more training seems to focus the embedding to more distinct topics and tends to remove synonyms / emoji with multiple semantics.
    Thus, depending on the application, both variants may be desired and fine-tuned.

    \afblock{Summary.}
    Our results show that Emoji2Emoji embeddings on social media posts reveal insightful associations that go beyond semantic groups in the Unicode standard.
    We posit that these associations are useful, \eg{} to understand the usage of each emoji in a given social media platform (\eg{} for its users).

    \begin{table*}[t]
        \centering
        \footnotesize
        \begin{tabular}{c|c|l}
            Class & Emoji & Top matching words (left to right)  \\ \hline\hline
\multirow{2}{*}{single} & \multirow{2}{*}{\emoji{emoji_u1f37b}} & \pbox{\linewidth}{\scriptsize bier, prost, nächst, mal, mein, gut, erstmal, letzt, schön, abend} \\ 
&& \textit{beer, cheers, next time, my, well, first, last, fine, evening}\\ \hline 
\multirow{2}{*}{single} & \multirow{2}{*}{\emoji{emoji_u1f3d5}} & \pbox{\linewidth}{\scriptsize camping, campen, zelten, hurricane, fernweh, festivals, reisetipps, verreisen, bereisen, urlaubsziel} \\ 
&& \textit{camping, camping (v), tents, (festival), wanderlust, festivals, travel, travel tips, traveling, visit, travel destination}\\ \hline 
\multirow{2}{*}{single} & \multirow{2}{*}{\emoji{emoji_u1f575}} & \pbox{\linewidth}{\scriptsize enttarnen, verdächtigen, entlarven, undercover, auffällig, sherlock, mysteriös, ermitteln, handeln, geheime} \\ 
&& \textit{expose, suspect, unmask, undercover, conspicuous, sherlock, mysterious, investigate, act, secret}\\ \hline 
\multirow{2}{*}{single} & \multirow{2}{*}{\emoji{emoji_u1f308}} & \pbox{\linewidth}{\scriptsize all, schön, the, gleichzeitig, sogar, bestimmen, and, bunt, ehe, gerne} \\ 
&& \textit{all, beautiful, the, simultaneous, even, determine, and, colorful, marriage, gladly}\\ \hline 
\multirow{2}{*}{single*} & \multirow{2}{*}{\emoji{emoji_u1f351}} & \pbox{\linewidth}{\scriptsize booty, butt, po, hintern, boobs, ass, brüste, tanga, datass, brüste} \\ 
&& \textit{(buttocks coll.), (breasts coll.), thong, (buttocks coll.), breasts}\\ \hline 
\multirow{2}{*}{multi} & \multirow{2}{*}{\emoji{emoji_u1f528}} & \pbox{\linewidth}{\scriptsize hammer, hämmern, hörmalwerdahämmert, lörres, presslufthammer, dlrh, werkzeug, vorschlaghammer, nageln, ding} \\ 
&& \textit{hammer, hammering, (TV show), (penis), pneumatic hammer, (coitus, platform), tool, sledgehammer, nail, thing}\\ \hline 
\multirow{2}{*}{multi} & \multirow{2}{*}{\emoji{emoji_u1f3f0}} & \pbox{\linewidth}{\scriptsize burg, hogwarts, ilvermorny, heidelberg, schloss, prinz, königreich, insel, emojis, beasts} \\ 
&& \textit{castle, hogwarts, ilvermorny, heidelberg, palace, prince, kingdom, island, emojis, beasts}\\ \hline 
\multirow{2}{*}{omni} & \multirow{2}{*}{\emoji{emoji_u1f648}} & \pbox{\linewidth}{\scriptsize hab, mal, weiß, einfach, sehen, eigentlich, fragen, echt, denken, grad} \\ 
&& \textit{just, sometimes, white, just, see, actually, ask, really, think, just}\\ \hline 
\multirow{2}{*}{omni} & \multirow{2}{*}{\emoji{emoji_u2795}} & \pbox{\linewidth}{\scriptsize zusammenfügen, vollständig, erhalten, einstellungen, aktuell, auswählen, klickt, siehe, vorhanden, hinzufügen} \\ 
&& \textit{merge, complete, get, settings, current, select, click, see, available, add}\\ \hline 
\multirow{2}{*}{omni} & \multirow{2}{*}{\emoji{emoji_u1f609}} & \pbox{\linewidth}{\scriptsize spaß, gerne, bitte, falls, suche, ps, suchen, danke, schön, zumindest} \\ 
&& \textit{fun, gladly, please, if, search, ps, search, thanks, beautiful, at least}\\ \hline 
\multirow{2}{*}{platform} & \multirow{2}{*}{\emoji{emoji_u1f68c}} & \pbox{\linewidth}{\scriptsize manni, busfahrer, racingteam, mannis, linie, kvgracingteam, busse, racing, hvvracingteam, formelaseag} \\ 
&& \textit{(name), bus driver, racing team, (name), bus line, (local transportation), busses, racing, (local transportation) }\\ \hline

        \end{tabular}
        \caption{
            \textbf{Emoji2Text association examples.}
            This table shows the top 10 words to a set of emojis filtered by a minimum document frequency of 500 in our dataset.
            We show the original German words and an English translation in italics below.
            This hand-picked selection aims to cover a broad range of emoji in four {\em classes} of emoji semantics: {\em i)} single-semantic emojis, {\em ii)} multi-semantic emojis, {\em iii)} omni-semantic emojis where the associations do not make an apparent sense, and {\em iv)} emojis associated with platform-specific idioms.
            The resulting word lists often provide a good textual emoji representation, whereas multiple semantics may mix.
        }
        \label{tab:emoji2text_similarity_example}
    \end{table*}

\section{Emoji2Text Associations}
\label{sec:E2T}
    Next, we use our embedding to associate Emoji2Text.
    One use case of this association is to improve keyboard predictions on mobile devices.
    Another one is to provide social network users a better understanding of the meaning of emojis in the target social media platform, by explaining emoji with words---which can be more descriptive the previously presented Emoji2Emoji associations.
    
    To give a first insight into the Emoji2Text associations, we study the top matching words for a set of emoji from different groups.
    Our first observation is that some of the top matching words are very specific to the Jodel platform, \eg{} \mbox{\emoji{emoji_u1f385} $\rightarrow{}${\em roterweihnachtsjodel}} (red x-mas Jodel), or \mbox{\emoji{emoji_u26bd} $\rightarrow{}${\em grünerfußballjodel}} (green soccer Jodel).
    This reflects specific user habits of posting a word/noun as a Jodel post in relation to an underlying in-app post background color.
    Evaluated by ourselves, other words can often be semantically linked to the specific emoji, whereas some cannot.
    
    To generalize and improve results, we filtered the resulting words by their document frequency.
    By adjusting this value to 500, we achieve better matchings according to our interpretation, yet we still find platform-specific and composite words within these sets as they occur quite often.
    Some emojis are used in a more specific sense than others, which we want to clarify by giving hand-picked examples in Table~\ref{tab:emoji2text_similarity_example}.
    This table shows an emoji with its Top 10 words (frequency greater than 500) that are closest within the embedding according to the cosine vector distance.
    While we provide the actual German words, we also give translations below each of these words.
    
    To perform a first qualitative Emoji2Text analysis, we introduce 4 \emph{exemplary} emoji classes based on their semantic variability in the top 10 words and their Emoji2Emoji association as determined by us:
    {\em i) single semantic emojis} have only a single association, {\em ii) multi-semantic emojis} have multiple associations (meanings), {\em iii) omni-emojis} having no specific apparent semantic, and {\em iv) platform-specific semantic emojis} have a Jodel-specific meaning that does not exist outside of Jodel.
    Next, we will discuss an explain the given example and our choice of class.
    
    \afblock{Single semantic emojis.}
    For this class, we found only semantically matching top 10 words within our embedding.
    That is, \emoji{emoji_u1f37b} refers to drinking, alcohol and beer, \emoji{emoji_u1f3d5} relates to camping, festivals and traveling.
    \emoji{emoji_u1f575} relates to suspicion and detecting and related instances, such as Sherlock Holmes.
    Interestingly, \emoji{emoji_u1f351} does not relate to its sense as a fruit as given by the Unicode definition within the top 10 words but is associated with various colloquial terms for buttocks.
    This phenomenon is not specific to the Jodel, but an established synonymous usage in personal digital communication.

    \afblock{Multi semantic emojis.}
    Other emojis have multiple semantics, \eg{} \emoji{emoji_u1f308} is associated with gay pride and LGBT, whereas it naturally also simply describes rainbows.
    For \emoji{emoji_u1f3f0}, we observe castles and kingdoms, the city of Heidelberg, and it also relates to Harry Potter.
    Looking deeper into this particular example, we also find references to the shire from LOTR.
    The \emoji{emoji_u1f528} relates to other tools, but may also have a colloquial different interpretation.

    \afblock{Omni semantic emojis.}
    Other emojis do not convey specific semantic associations, such as the symbol \emoji{emoji_u2795}.
    The example of \emoji{emoji_u1f648} is associated with embarrassment, awkwardness and weirdness.
    That is, it may be used in various (also possibly platform-specific) contexts.
    Quite generic emotion emojis experience heavy usage resulting in high frequencies within our dataset.
    Finally, \emoji{emoji_u1f609} shows that there is no apparent semantic linked to it except for positivity represented by all face-positive emojis.
    
    \afblock{platform-specific emoji semantics.}
    Some emojis develop a special semantic within a platform, which is reflected in our embedding.
    A good example is \emoji{emoji_u1f68c}.
    The top 10 words refer to busses, the German forename ``manni'', and different public transportation providers as a meme referring to their service quality (\eg{} ``racing team'' to reflect slow running busses): here, we find the local service corporations KVG (Cologne), ASEAG (Aachen) and HVV (Hamburg) linked to the mentioned meme.
    The name ``manni'' is platform-specifically used as a synonym for bus drivers.

    \afblock{Summary \& Limitations.}
    By showcasing emoji from a broad set of different groups, we find strong evidence for good semantic associations between emojis and words within the embedding.
    Although some emojis may inherently not convey particular semantics, most do, which is reflected within the given examples.
    We also find multiple semantics for a given emoji due to lacking capabilities for any context in such classical word embeddings.
    Note that the preliminary distinction between the introduced classes is not always straight forward and limited to our interpretation.
    While this is a first look into the Emoji2Text associations, a broader evaluation incorporating human judgment would be the next step, which we leave for future work.

\section{Text2Emoji Associations}
    As the last application, we aim at using embeddings to associate emojis to a given word (Text2Emoji).
    This association likely can be used in several applications such as giving users predictions of emojis while writing or to translate text to emoji.
    Yet again, these predictions might also help choosing a suitable emoji, given platform particularities.
    Further, they may help understanding the perceived meaning a text may convey within a specific community.
    
    We decided to use a quantitative analysis for the Text2Emoji associations to give more variance in presenting possible applications.
    To evaluate the applicability of leveraging word embeddings for the Text2Emoji association, we first define our target task.
    For keeping this downstream task simple, we define this task as {\em predicting the first occurring emoji within a given text} (disregarding others).
    Other target functions may be approached likewise.

    Our evaluation consists of two different approaches: {\em i)} Naive direct prediction directly using the word embedding, and {\em ii)} applying statistical machine learning tools as a smart indirection layer on this problem.
    
    \afblock{Data preparation \& selection.}
    As described in Section {\em Approach: Word-Emoji Embeddings}, our data has been cleaned and lemmatized before doing any further processing. %
    
    To enable a viable evaluation, we first needed to balance our data since the emoji popularity is heavy-tailed.
    We selected all emojis whose overall share in usage is higher than $0.1\permille{}$.
    This leaves us with  117 different emojis each having a first occurrence in at least $11.5\si{k}$ of distinct sentences.
    From our dataset, we then randomly selected this number of sentences per emoji that match our problem definition of the selected emoji being the first one occurring in each sentence.
    These emoji sub-datasets ($\approx 1.3\si{M}$ sentences) are then split into 5 partitions ($\approx 270\si{k}$ sentences) enabling a 5-fold cross validation in our ML approach.
    
    \afblock{Test setup.}
    For both evaluation approaches, we create a {\em base} word embedding from all previously non-selected sentences ($\approx 40.9\si{M}$) masking any emoji that might occur, such that this embedding only contains words.
    For all embeddings, we used the word2vec implementation with 300 dimensions.
    Then, for each fold, we individually continue training on the base embedding with 4 out of 5 emoji data subsets resulting in 5 different embeddings---each excluding a single subset that is later on used for validation.
    
    \afblock{Feature selection.}
    To generate features from an input sentence $s$, we mask all non-first occurring emojis and then calculate an aggregate over all word vectors $v\in V_{s'}$ from the used embedding as proposed in~\cite{de2016representation}: $f_{s'}=\text{mean}\left(v(w), w \in s'\right)$.

    For our dataset, the mean performs slightly better than the median for the naive approach, whereas min or max achieve worse results with a higher variance; therefore, we decided to use a mean aggregation.
    
    \afblock{Learning methods.}
    We apply two methodically different approaches to our prediction task.
    {\em i)} We implemented a naive method that calculates the top $k$ most similar emojis directly within the word embedding by the cosine distance.
    In this case, there is no need to train an additional indirection layer, \ie{} for each of the 5 embeddings, we can evaluate the non-matching other 4 subsets.
    {\em ii)} We further applied a set of commonly available machine learning techniques (\eg{} RandomForest, LogisticRegression) with a limited set of hyperparameters.
    For all of the latter, we created 5 training and validation sets, while randomly shuffling the training set between different folds.
    We used the resulting probability matrix for calculating the top k predictions.
    While development, we noticed that training and validation on only a small subset provide quite similar results.
    Yet for this paper, we present only results on the full training and test set.

    \subsection{Results}
    \afblock{Baseline.}
    A classical ZeroR baseline would choose the class that is predominant within the set, however, due to our balanced data subset, chances are equal.
    Thus, the probability in a first try is $1/|\text{classes}|$, which is $\approx 0.0085$ for our $n=117$ classes.
    In case we are having multiple consecutive guesses, we intuitively compute the probability as $p(n,k)=p(n, k-1)+(1-p(n, k-1)) \cdot (n-k)^{-1},\ p(n, 0)=n^{-1}$.

    \begin{table}[t]
        \centering
        \footnotesize
        \resizebox{\linewidth}{!}{        
            \begin{tabular}{l|c|c|c|c|c}
                Method & Top1 & Top2 & Top3 & Top4 & Top5 \\ \hline\hline
                ZeroR  & $0.0085$ & $0.0171$ & $0.0256$ & $0.0342$ & $0.0427$ \\ \hline\hline
                Naive  & $0.0847$ & $0.1340$ & $0.1679$ & $0.1964$ & $0.2208$ \\ 
                       & $\pm0.0014$ & $\pm0.00125$ & $\pm0.0011$ & $\pm0.0011$ & $\pm0.0012$ \\ \hline\hline
                MLPerceptron    & $\textbf{0.1292}$    & $\textbf{0.1932}$    & $\textbf{0.2335}$    & $\textbf{0.2660}$    & $\textbf{0.2935}$ \\
                                & $\pm0.0001$ & $\pm0.0004$ & $\pm0.0005$ & $\pm0.0072$ & $\pm0.0008$ \\ \hline
                LogRegression   & $0.1221$    & $0.1828$    & $0.2224$    & $0.2545$    & $0.2821$ \\
                                & $\pm0.0008$ & $\pm0.0005$ & $\pm0.0006$ & $\pm0.0004$ & $\pm0.0005$ \\ \hline
                RandForest      & $0.1140$    & $0.1645$    & $0.1972$    & $0.2241$    & $0.2483$ \\
                                & $\pm0.0005$ & $\pm0.0003$ & $\pm0.0005$ & $\pm0.0006$ & $\pm0.0008$ \\ \hline
                AdaBoost        & $0.0926$    & $0.1437$    & $0.1792$    & $0.2079$    & $0.2323$ \\
                                & $\pm0.0014$ & $\pm0.0001$ & $\pm0.0009$ & $\pm0.0007$ & $\pm0.0006$ \\ \hline
                GaussianNB      & $0.0797$    & $0.1207$    & $0.1490$    & $0.1731$    & $0.1943$ \\
                                & $\pm0.0003$ & $\pm0.0004$ & $\pm0.0002$ & $\pm0.0006$ & $\pm0.0008$ \\ \hline
            \end{tabular}
        }
        \caption{
            \textbf{Text2Emoji prediction results.}
            This table summarizes the first occurring emoji prediction precision results for the top 1 to 5.
            The ZeroR baseline depicts random choice, our naive approach directly uses the embedding, whereas others apply another indirection layer of machine learning.
            While the naive approach outperforms random choice, additional machine learning {\em can} significantly improves results.
            The best performing algorithm was the MPL being slightly better than LogisticRegression and RandomForest, whereas others compete with our naive approach.
            Most results are consistent across all top k predictions.
        }
        \label{tab:results_topk}
    \end{table}

    \afblock{Naive approach.}
    Our first attempt on directly matching a best suited emoji for sentences directly within the embedding by the cosine distance yields an accuracy of about $8.47\pm.14\%$ for an exact prediction as shown in Table~\ref{tab:results_topk} (Top 1 column, second row).
    Comparing this result to the baseline (ZeroR, first row), our naive approach performs an order of magnitude better.
    By loosening the problem allowing a set of ranked predictions, we also show further Top 2..5 results.
    Here, the delta to the baseline gets smaller for the top k predictions.
    Presumably due to our large dataset, the standard deviation across the folding sets is quite small.
    The accuracy of the top $k$ predictions increases almost linearly, such that we can predict the first emoji within the top 5 set with a precision of $22.08 \pm~0.12\%$.
    
\begin{figure*}[t]
        \label{fig:confusion_ml}
        \begin{subfigure}{.45\linewidth}
            \includegraphics[width=\linewidth]{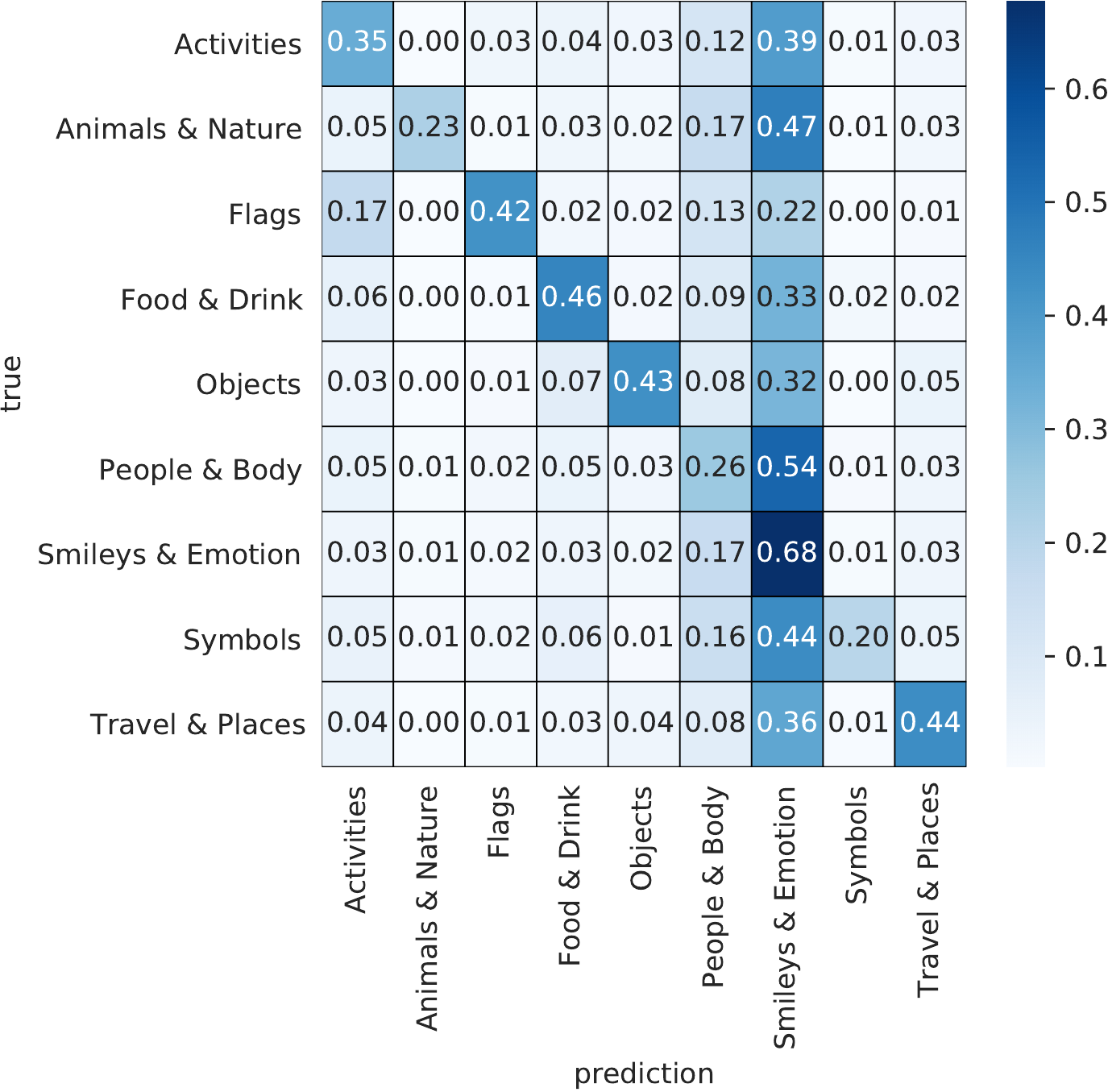}
            \subcaption{Naive approach}
            \label{fig:result_ml_naive_confusion}
        \end{subfigure}
        \qquad
        \begin{subfigure}{.45\linewidth}
            \includegraphics[width=\linewidth]{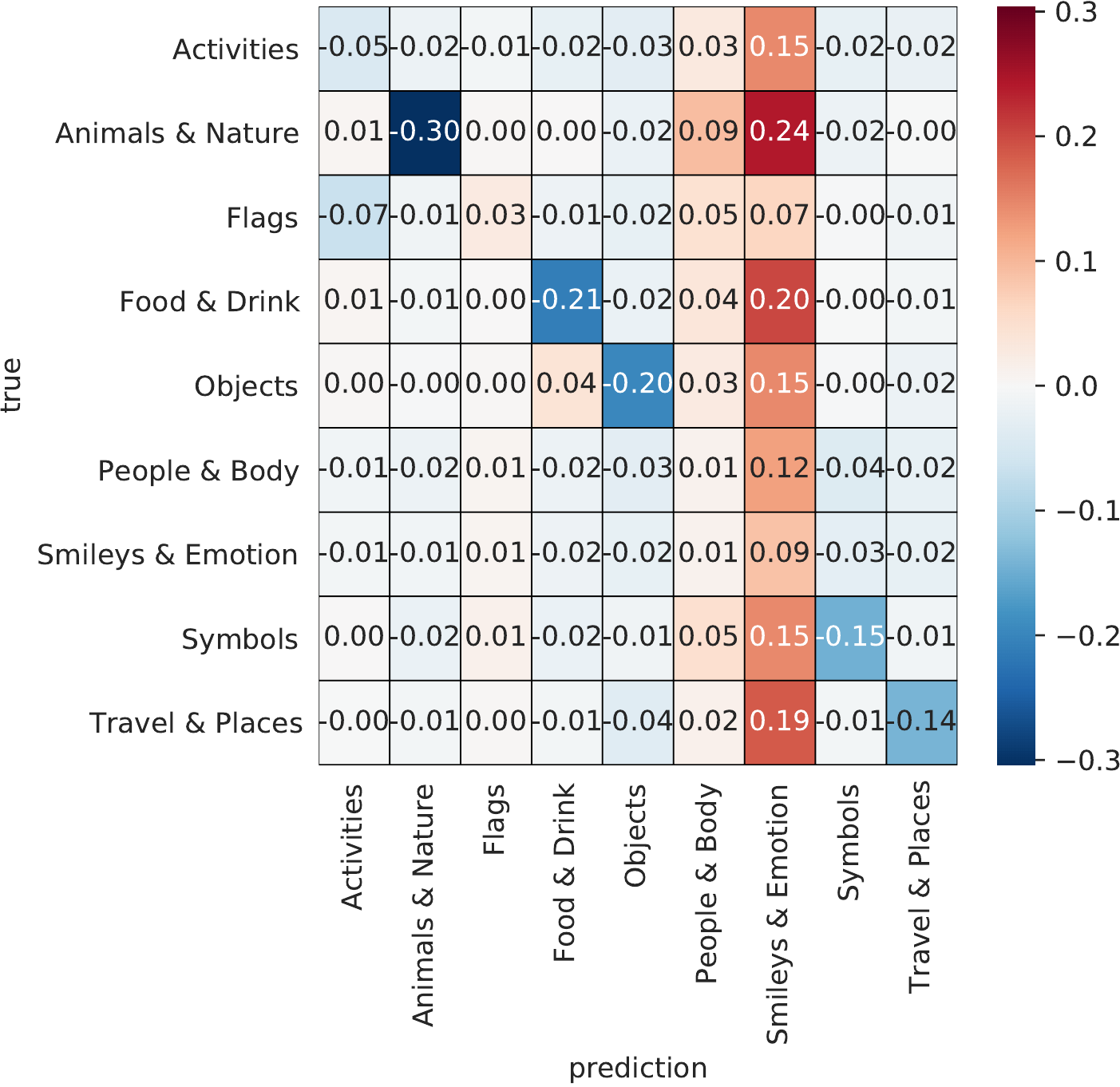}
            \subcaption{Difference: naive approach - machine learning}
            \label{fig:result_ml_confusion}
        \end{subfigure}
        \caption{
            \textbf{Text2Emoji top 1 prediction confusion matrices by Unicode groups.}
            The given heatmaps depict the confusion between the Unicode emoji group for the top 1 prediction task.
            Each actual emoji group (y-axis, true) is mapped to a group of a predicted emoji (x-axis, prediction).
            \textbf{Subfigure~\ref{fig:result_ml_naive_confusion}} shows the prediction confusion for the naive approach (normed row-wise to $1$).
            We observe a diagonal that reveals mostly a good match, however, there is a strong tendency towards predicting emojis from the group {\em Smileys \& Emotion}.
            \textbf{Subfigure~\ref{fig:result_ml_confusion}} depicts the same analogy, but shows the difference from the naive approach to the best performing machine learning variant.
            Values below $0$ imply a higher value in comparison; values greater $0$ likewise a lower value.
            \Eg{} the predictions towards {\em Smileys \& Emotion} are consistently less frequent.
            As the ML approach performs better, the difference is almost consistently negative on the diagonal, whereas other values are mostly positive.
        }
    \end{figure*}

    To get a better insight into where this algorithm fails in particular, we also analyzed the resulting confusion matrix shown in Figure~\ref{fig:result_ml_naive_confusion}.
    This Figure shows the emoji group of the actual--true--first emoji of a sentence (y-axis) in relation to the predicted emoji's group (x-axis).
    We normed the values, such that each row sums up to $1$.
    A perfect match would result in a straight diagonal, whereas a random prediction would yield an equal distribution across the heatmap.
        
    Although many predictions seem reasonable on an emoji group level, mispredictions predominantly towards {\em Smileys \& Emotion}, and {\em People \& Body}, are apparent.
    On a deeper level---the confusion between subgroups---we observe the same picture, in particular a shift towards {\em Face-Concerned}, {\em Face-Smiling} and {\em Hand-Signs} (not shown).
    Although this needs to be analyzed deeper in the future, we believe that this is a result of the applicability and usage of these emojis in many different contexts.
    Further, the grouping defined by the Unicode standard may not be optimal in a semantic sense as shown by~\cite{barbieri2016does}, who propose using data-driven clustering of emojis.

    \afblock{Machine Learning.}
    Secondly, we applied off the shelf machine learning techniques to the problem of predicting the first occurring emoji within a text.
    Our choice of implementation was the Python sklearn package due to its simplicity and popularity.
    This allows us to plug in our task into various algorithms easily.
    Our set of choices consists of GaussianNB, AdaBoost, RandForst, LogRegression, and MLP.

    We ran a grid search over several hyperparameter combinations of which we only show the best results for each algorithm in Table~\ref{tab:results_topk}.
    This table shows the accuracy of prediction in two different perspectives: {\em i)} the top 1 column depicts the resulting prediction of each algorithm, whereas {\em 2)} the top 2..5 columns depict the accuracy of each algorithm according to its resulting probability table.
    That is, whether a correct prediction was amongst the $k$ most probable results.
    Each value is represented by the mean and standard deviation across the 5 folds.
    
    All classifiers outperformed the ZeroR baseline clearly.
    The best performing algorithm was Multi-Layer Perceptron classifier with a top 1 (5) accuracy of $12.9\%$ ($29.4\%$), yet the achieved precision is very close to LogisticRegression and RandomForest throughout any top k prediction.
    Other classifiers performed worse and are comparably good to our naive approach.

    As seen in the naive approach, the prediction task is hard for specific groups of emojis (cf. Figure~\ref{fig:result_ml_naive_confusion}).
    We also present these metrics of our best performing machine learning approach in difference to the naive approach in Figure~\ref{fig:result_ml_confusion}.
    \Eg{} the predictions towards {\em Smileys \& Emotion} are consistently less frequent.
    In comparison, the better performing machine learning algorithm provides better results for the predominantly mispredicted groups as the difference is almost consistently negative on the diagonal, whereas other values are mostly positive.
    Still, we draw the same conclusion: there is a shift towards {\em Smileys \& Emotion} and {\em People \& Body} on a group level.
    In particular, the subgroup confusion matrix reveals that commonly used emotion and hand-sign emojis are the main drivers of mispredictions (not shown).

    \afblock{Summary.}
    Machine learning can be used to predict an emoji for a given word (text to emoji translation) that improves accuracy over the naive approach.
    While some machine learning techniques perform better than the naive approach, they need an additional considerable amount of computing power for training.

    An intense grid search over the hyperparameter space might provide better results.
    However, since the aim of our study is to demonstrate feasibility, we leave this task open for future work.

\section{The JEED1488 Dataset}
\label{sec:jeed1500}
    We release the resulting emoji to emoji embedding data set, consisting of 1488 keyed vectors for each used emoji in a 300-dimensional space based on about $42\si{M}$ posts from the Jodel messaging app as described in Section {\em Approach: Word-Emoji Embeddings}. %
    Due to privacy concerns, we cannot publish the full embedding, yet we publish the emoji sub-embedding. %
    That is, provided files contain 1488 emoji-to-vector entries.
    We demonstrate the basic usage in additional scripts.
    It can obtained from GitHub\footnote{\url{https://github.com/tuedelue/jeed}}.
    Within this paper, we have shown its applicability for basic use-cases, yet it will allow and should encourage further research.

\section{Related Work}
\label{sec:related_work}

    Emojis are widely studied.
    To mention a few, empirical measures on emoji usage~\cite{lu2016learning,ljubevsic2016global}, influences of context and culture~\cite{takahashi2017smiling}, expression \& perception~\cite{li2019exploring,berengueres2017differences} and possibly misinterpretations~\cite{miller2016blissfully} or irony~\cite{gonzalez2011identifying}.
    Another topic is sentiment analysis on social networks that often is performed on a word level, but has also attracted incorporating emojis~\cite{hu2017spice,novak2015sentiment}.

    \afblock{Emoji2Emoji.}
    The Emoji2Emoji task was analyzed on \eg{} Twitter data~\cite{illendula2018learning,barbieri2016does}.
    Barbieri et al. define two questions: {\em i)} topical similarity---do two emojis occur at the same time? And {\em ii)} functional similarity---can two emojis be used interchangeably?
    Its evaluation leverages human judgment for both questions of 50 selected emoji pairs and achieves an accuracy of $78\%$ for both tasks.
    A qualitative evaluation is given by a clustered t-SNE evaluation.
    In \cite{wijeratne2017semantics}, Wijneratne et al. extend their prior work on a knowledge database of emoji semantics~\cite{wijeratne2017emojinet}.
    They add {\em sense} information into an emoji embedding, which is validated on 508 emoji pairs in terms of {\em i)} and {\em ii)} via crowdsourcing.
    Their model mostly is strongly correlated to human judgment.
    As they have published the annotation results, we find our used embedding instance providing similarities only moderately in line with human judgment.
    
    \afblock{Emoji2Text.}
    \cite{eisner-etal-2016-emoji2vec} use textual descriptions of emojis to map them into the Google News Dataset word embedding.
    By doing so, they obtained 85\% accuracy in predicting keywords for manually labeled emojis. 
    An emoji knowledge base extracted from Google News and Twitter data including sense labels has been presented in~\cite{wijeratne2017emojinet}.
    Within a multi-staged crowdsourced human evaluation, they show $83.5\%$ valid sense assignments.

    \afblock{Text2Emoji.}
    \Eg{} \cite{felbo-etal-2017-using} create a word-emoji embedding using LSTM on Twitter data.
    They showcase a Text2Emoji downstream task predicting one out of 64 emojis by applying deep learning and achieve a Top 1 (5) accuracy of $17.0\%$ ($43.8\%$) outperforming fasttext-only which yielded accuracy values of Top 1 (5) of $12.8\%$ ($36.2\%$) on a balanced subset.
    Another work~\cite{zhao2018analyzing} use a multi-modal deep learning model that predicts an emoji and its position within the text also leveraging image and user demographic information.
    This approach predicts the correct emoji out of 35 most frequent emojis with a Top 1 (5) accuracy of $38\%$ ($65\%$).
    Further, \eg{} \cite{barbieri2017emojis} used an LSTM for prediction, whereas \cite{guibon2018emoji} use tf-idf and argue that an enrichment with sentiment features improves prediction quality.
    
    We complement related work by performing all 3 tasks to study real-world semantic associations on a large-scale messaging data.

\section{Conclusions \& Future Work}
    We showed that embeddings are useful to study real-world semantic associations of emoji and words on large-scale messaging data.
    Our results indicate that word-emoji embeddings reveal insightful Emoji2Emoji and Emoji2Text associations on social media posts going beyond semantic groups defined by the Unicode standard.
    We show that emoji prediction directly from the embedding may work reasonable well; however, machine learning can improve the results significantly.
    We posit that such associations are key to understand the usage of each emoji in a given social media platform (\eg{} for its users).
    
    While our work demonstrates the potential usefulness of Word-Emoji embeddings for large scale messaging data, it is exploratory and uses qualitative inspections as a major instrument for our investigations. 
    We used Word2Vec for creating the embedding, other embedding approaches like FastText~\cite{bojanowski2017enriching}.
    Leveraging context, specifically trained sentence-embedding models~\cite{kenter2016siamese}, or, \eg{} Bert~\cite{devlin2018bert}, may further improve results and possibly catch multiple semantics better.
    Further, it is still unclear how exactly the amount of training influences semantics that can be extracted from the embedding; we find that more is not always better, depending on the desired application.
    Besides technical improvements, future work should explore real-world semantic associations in a more principled manner by, \eg{} incorporating human subjects as evaluators or annotators. 
    The Jodel app has the unique features of being location-based and anonymous.
    This enables interesting questions of sorting out semantic differences between locations; the anonymity may also introduce specific semantic aspects different to other Social Media.
    
    We enable such investigations partly by releasing the JEED1488 emoji-subembedding with the publication of this paper, and hope to inspire more research into emoji related downstream tasks.

\bibliographystyle{aaai} 
\bibliography{samplebib}

\end{document}